\documentclass{article} 
\usepackage{iclr2023_conference_tinypaper,times}

\usepackage{amsmath,amsfonts,bm}

\def\eqref#1{equation~\ref{#1}}

\def\1{\bm{1}}

\DeclareMathAlphabet{\mathsfit}{\encodingdefault}{\sfdefault}{m}{sl}
\SetMathAlphabet{\mathsfit}{bold}{\encodingdefault}{\sfdefault}{bx}{n}

\usepackage{hyperref}
\usepackage{url}
\usepackage{graphicx}
\usepackage{multirow}
\usepackage{booktabs}
\usepackage{caption}
\usepackage{wrapfig}
\usepackage{array}
\usepackage{wasysym}

\title{Collapse of Self-trained Language Models}

\author{David Herel \\
FEE, Czech Technical University in Prague \\
\texttt{hereldav@fel.cvut.cz} \\
\And
Tomas Mikolov\\
CIIRC, Czech Technical University in Prague \\
\texttt{tmikolov@gmail.com}
}

\iclrfinalcopy 
\begin{document}

\maketitle

\begin{abstract}
In various fields of knowledge creation, including science, new ideas often build on pre-existing information. In this work, we explore this concept within the context of language models. Specifically, we explore the potential of self-training models on their own outputs, akin to how humans learn and build on their previous thoughts and actions. While this approach is intuitively appealing, our research reveals its practical limitations. We find that extended self-training of the GPT-2 model leads to a significant degradation in performance, resulting in repetitive and collapsed token output.
\end{abstract}

\section{Introduction \& Related Work}
\label{intro}
\vspace{-3pt}
From the viewpoint of artificial intelligence, it could be important for a model to be able to self-evolve and learn from its own actions. Although neural network models partially address this problem by storing information in the hidden layer and utilizing the attention mechanism, for example, the vanishing gradient problem limits their effectiveness \citep{vanishing}. Dynamic models \citep{jelinek-etal-1991-dynamic} have been suggested as a solution where models are trained on test data to utilize a form of cache. Dynamic evaluation for neural networks models was proposed by \cite{mikolov-dynamic, mikolov-dynamic-2, kraus-dynamic, kraus-transformers}, where the neural network parameters are updated using the standard training mechanism during the processing of the test data.

Our work explores the concept of self-training a model on its own output that is generated through sampling \citep{deoras-gen}. However, we provide empirical evidence that this self-training approach can lead to model collapse, where the generated outputs become severely biased and repetitive. This trend has also been explored in \cite{shumailov2023curse}. Our findings indicate limitations in the current model architecture regarding self-evolution. For future research, it may be beneficial to explore entirely new models that can more effectively accommodate this aspect.

\section{Method: Self-training of LLM}
\vspace{-5pt}
In our settings, self-training adjusts model parameters $\theta_g$ to better model local sequence distribution, $P_l(x)$, which is generated from a model. The initial adapted parameters $\theta0_l$ are set to $\theta_g$, computing the probability of the first sequence, $P(s_1|\theta0_l)$. This results in a cross-entropy loss $\mathcal{L}(s_1)$, with gradient $\nabla\mathcal{L}(s_1)$, updating the model to adapted parameters $\theta1_l$. We then let the model generate sequence again and evaluate $P(s_2|\theta1_l)$, repeating this for each $s_i$. Each update approximates the current local distribution $P_l(x)$.
\vspace{-5pt}
\begin{figure}[h]
\centering
\includegraphics[width=0.55\textwidth]{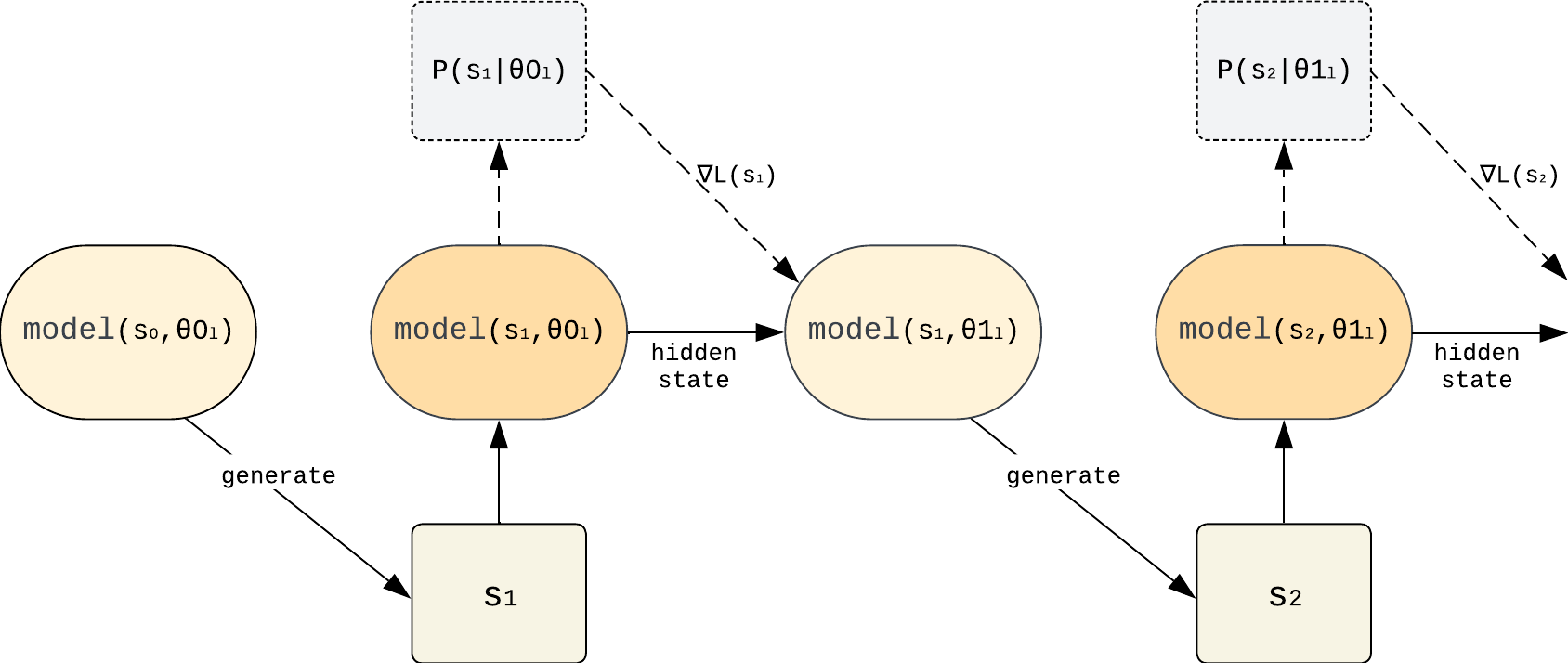}
\vspace{-5pt}
\caption{Schema of the self-training. The model generates a sequence $s_1$, computes its probability $P(s_1|\theta0_l)$, which is then used to determine the cross-entropy loss with gradient $\nabla\mathcal{L}(s_1)$ to update the next state of the model with the adapted parameters.}
\end{figure}

\section{Experiment: Empirical analysis of GPT-2 Model} 
For our experiments, we utilized the pre-trained GPT-2 \citep{gpt-2} model that is available as open-source. We allowed the model to train on its own generated output while tracking its performance after each update on a valid set of Wikitext-2 \citep{wikitext-2}. We set a stopping criterion for the model, either when it collapsed to repeating sequences or when it reached 1000 iterations. The hyperparameters used in our experiments, as well as the associated codebase, are available in \ref{hyperparams}.

Our observations show that the validation loss increases with each iteration and is significantly influenced by the learning rate. When the learning rate is higher, the model collapses faster and produces repetitive tokens quickly. This phenomenon is exemplified by a significant decrease in loss on generated (train) data, almost reaching 0 loss. The progression of output generation and the noticeable degradation towards model collapse can be observed in \ref{examples}. Further details on the impact of model size on the rate of collapse are provided in the appendix \ref{sizes}.

\begin{figure}[h]
\centering
\includegraphics[width=1\textwidth]{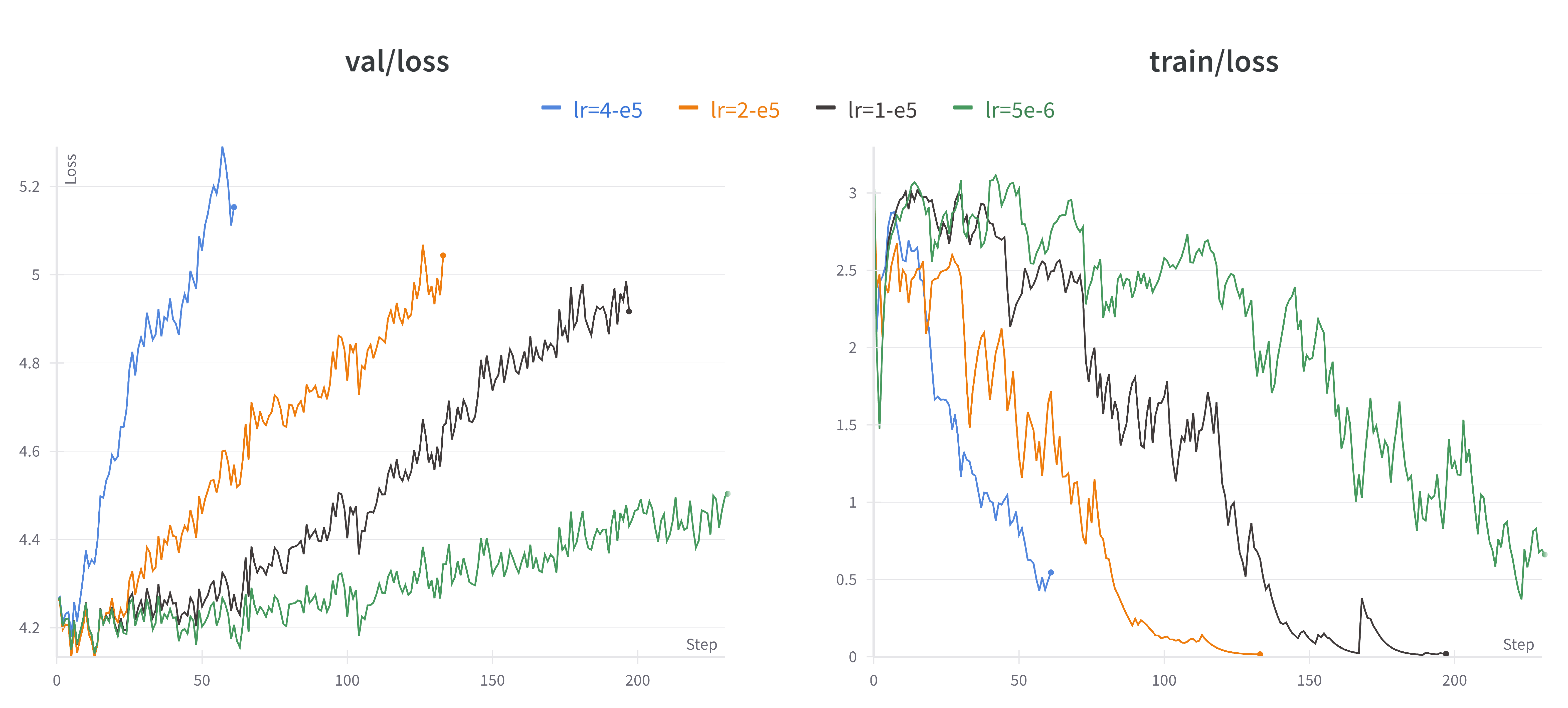}
\caption{Impact of learning rate on self-training GPT-2 \citep{gpt-2} language model on valid and train sets. As the learning rate increases, the model's performance deteriorates, leading to a higher loss on the valid set. On the train set, the model collapses and converges into a generation of repetitive tokens, resulting in almost zero loss on generated data. The y-axis represents the loss, and the x-axis displays the number of model steps.}
\end{figure}

\section{Discussion}
In this study, we investigated the potential of self-training language models on their own outputs. Our results demonstrate that extended self-training of the GPT-2 model leads to significant performance degradation, with models collapsing into repetitive sequences consistently. We also observe that the learning rate has a notable impact on the speed of this collapse.

With the extensive use of language models in various text generation applications, it can be expected that in the future there will be an increasing amount of text on the web with artificial origin. As the training data for language models are typically scraped from the web, the collapsing problem we describe in this paper can become a serious issue, as language models will be in the future largely trained on data that were generated from other language models.
\newpage
\subsubsection*{URM Statement}
The authors acknowledge that at least one key author of this work meets the URM criteria of ICLR 2024 Tiny Papers Track.

\bibliography{iclr2023_conference_tinypaper}
\bibliographystyle{iclr2023_conference_tinypaper}

\appendix
\section{Appendix}
\subsection{Example of model collapse}
\label{examples}
This section contains examples of sequences generated by the GPT-2 model with a learning rate of 2-e5. Starting from the first iteration, we provide a list of what the model generated at 50 iterations, 100 iterations, and the final iteration when the model collapsed into repetitive repetition, which is documented in Table \ref{table:examples}. These examples demonstrate the limitations of the current model architecture regarding self-evolution and highlight the challenges associated with self-training language models on their own outputs. By presenting these examples, we hope to provide valuable insights into the potential and challenges of self-training language models and to contribute to ongoing efforts to improve the performance and effectiveness of these models.

\begin{table*}[htbp]
    \centering
    \begin{tabular}{m{1.3cm}|m{12cm}}
     \toprule
     \bfseries Iteration & \bfseries Example \\
     \midrule
        0 & \textit{And these days, we can understand why so many entrepreneurs struggle with it. I've studied that fact thoroughly — along with many others. It is the reason we have the most value — not only at our company, but in other businesses around the world.} \\
        & \textit{We've got our share of challenges, too. Most of those we face are insurmountable. We're very lucky, even if our success holds a 1\% return.} \\
        & \textit{In America, 20 companies provide less than 1\% return, but this has dropped 3-4\% in 18 months. While the industry has increased, its margins are shrinking and its investment opportunities are curtailed.} \\
        & \textit{Take RFPs for some of the leading firms . Many startups fail because of weak leadership; some are successful, but still fail.} \\
        & \textit{High tech companies need competition. Take RFP for Google . The main competitor to RFP A is Data Corp .} \\
        & \textit{Sourcing remains a major choice for businesses, and so need} \\
     \midrule
        50 & \textit{The way to look at this is that the number of U.S. citizens in the "cannabis community" is not that large. And it can be if these are local residents, local or national.} \\
        & \textit{But at the same time the number of U.S. citizens in the cannabis community is not that large.} \\
        & \textit{The U.S. grows a lot of (cannabis)} \\
        & \textit{So we should be pretty sure that there is a lot of support.} \\
        & \textit{There's still a long way to go.} \\
        & \textit{But, I want to let you in on the good news.} \\
        & \textit{The federal law that will likely be up in the air in the coming months, before the House and Senate both vote on it, is the very same law that has already been introduced.} \\
        & \textit{The Affordable Care Act, those are the two main provisions.} \\
        & \textit{The law that I had to tell you, it's only so I can say that it's the} \\
     \midrule
        100 & \textit{- In the 4.1 final build, we dropped the 'the\_no\_go' flag.} \\
        & \textit{- In the 4.9 final build, we stuck the 'the\_no\_go' flag.} \\
        & \textit{- In the 4.9 final build, we stuck the 'the\_no\_go' flag.} \\
        & \textit{- In the 4.9 final build, we will be in the "final" release.} \\
        & \textit{- The "final" release is the "right time"} \\
        & \textit{- The "final" release is the "right time"} \\
        & \textit{- The "final" release has the "right time"} \\
        & \textit{- The "final" release has the "right time"} \\
        & \textit{- The "final" release is the "right time"} \\
        & \textit{- The "final" release is the "right time"} \\
        & \textit{- The "final" release is the "right"} \\
     \midrule
        133 (last) & \textit{-} \\
        & \textit{-} \\
        & \textit{-} \\
        & \textit{-} \\
        & \textit{-} \\
        & \textit{-} \\
        & \textit{-} \\
        & \textit{-} \\
        & \textit{-} \\
        & \textit{-} \\
        & \textit{-} \\
     \bottomrule
    \end{tabular}
    \caption{Examples at key iterations — 0, 50, 100, and the final iteration documenting the progression until the model succumbed to repetitive patterns.}
\label{table:examples}
\end{table*}

\newpage
\subsection{Impact of parameter sizes}
\label{sizes}
\begin{figure}[h]
\centering
\includegraphics[width=1\textwidth]{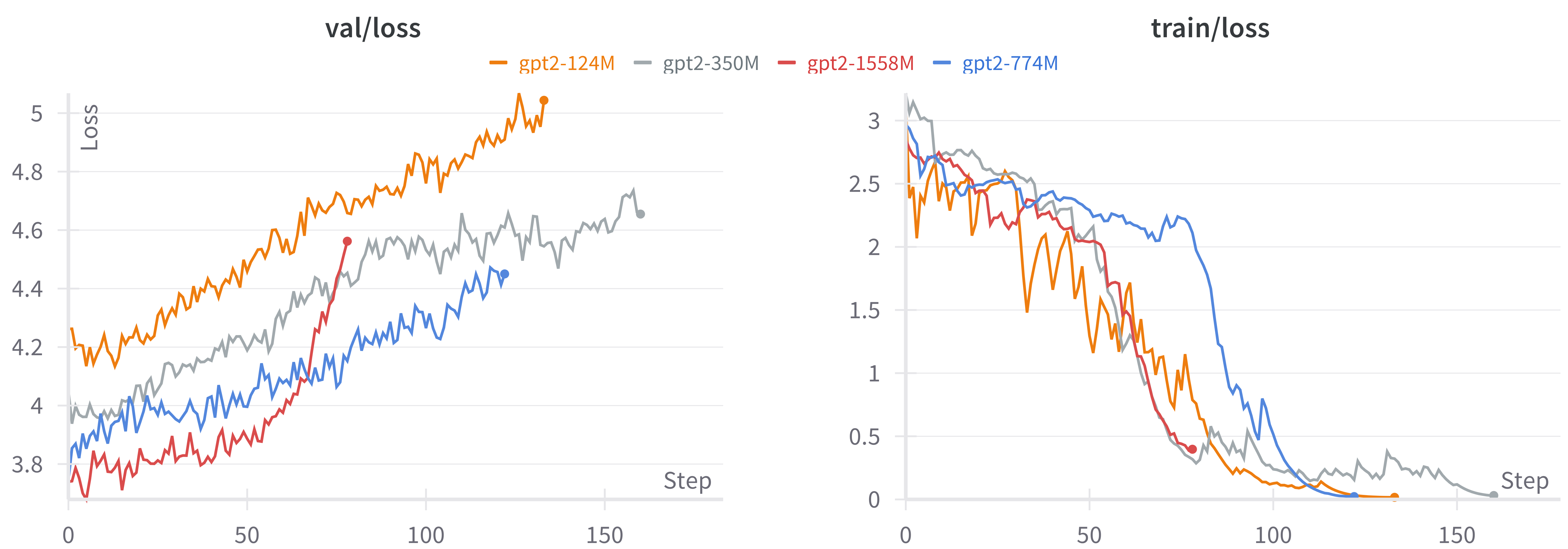}
\caption{Correlation between model size and the onset of collapse in GPT-2 architectures.}
\label{fig:sizes}
\end{figure}

To investigate the relationship between the number of parameters in a model and its stability, we conducted a series of experiments using GPT-2 architectures with varying sizes. Specifically, we compared models with parameter counts ranging from 100 million to 1.5 billion. As depicted in Figure \ref{fig:sizes}, our findings highlight a notable trend: larger models tend to exhibit more rapid onset of model collapse.

\subsection{Different evaluation dataset}
\label{ptb}
\begin{figure}[h]
\centering
\includegraphics[width=1\textwidth]{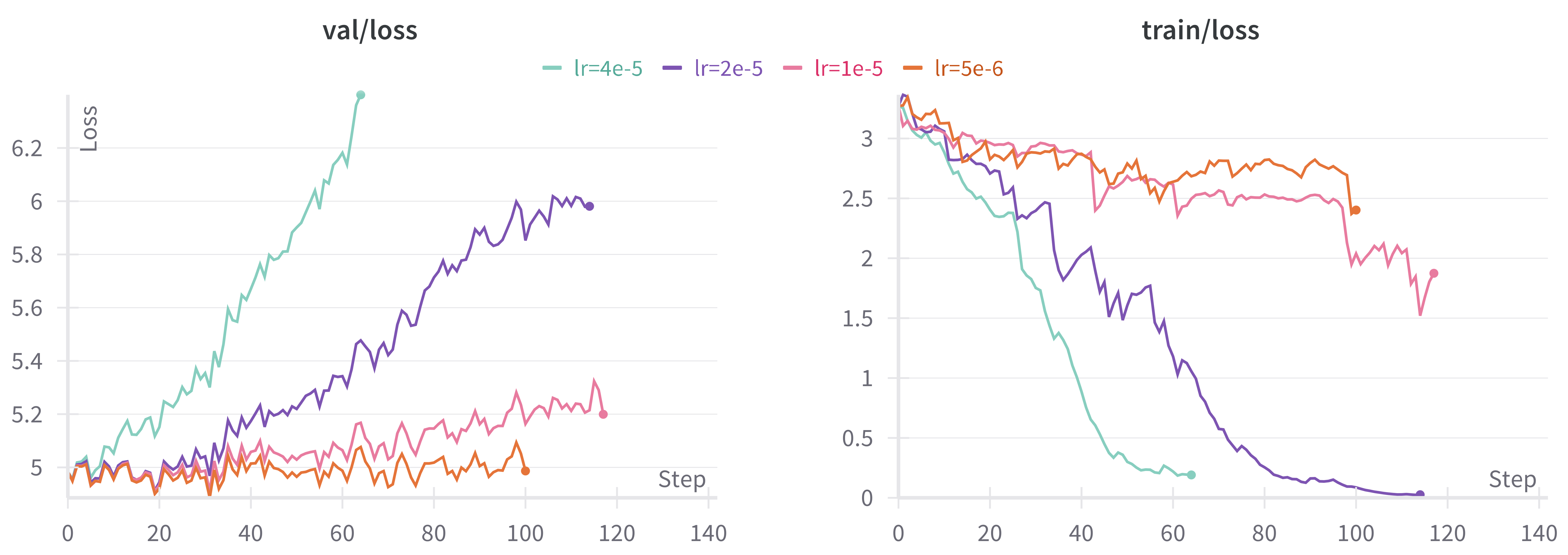}
\caption{Comparative analysis of learning rate impact on self-trained GPT-2 model performance, evaluated on it's output (train loss) and validation subset of the Penn Treebank dataset.}
\label{fig:ptb}
\end{figure}

Beyond the standard Wikitext-2 benchmark, we expanded our evaluation to include the Penn Treebank dataset \citep{FITPT283}, a prominent resource in language modeling. Our objective was to ascertain whether the increase in validation loss, as observed with the Wikitext-2 dataset, is consistent across different datasets. The comparative results presented in Figure \ref{fig:ptb} indicate that the Penn Treebank dataset yields a similar pattern, suggesting that our observations are not exclusive to a single dataset but may reflect a more general phenomenon.

\newpage
\subsection{Experiment hyper-parameters}
\label{hyperparams}
In this section, we present a list of the hyper-parameters utilized in our model experiments. Additionally, to promote transparency and reproducibility, we have created a code-base that replicates our results: \href{https://github.com/DavidHerel/collapse-lm-iclr/tree/main}{collapse-lm}

\begin{table}[h!]
\centering
\small
\begin{tabular*}{0.3\linewidth}{p{0.18\linewidth}|p{0.12\linewidth} }
 \hline
 Hyperparameters & Value \\
 \hline
 temperature & 0.8 \\
 top\_k & 500 \\
 max\_new\_tokens & 200  \\
 max\_iters  & 100 \\
 grad\_clip & 1 \\
 batch\_size & 1 \\
 block\_size & 100 \\
 n\_layer & 12  \\
 n\_head & 12  \\
 n\_embd  & 768 \\
 dropout & 0 \\
 bias & False \\
 beta1 & 0.9 \\
 beta2 & 0.95\\
 prompt & ""\\
 batch\_size & 1\\
 \hline
\end{tabular*}
\caption{Hyper-parameters for GPT-2 model experiments.}
\label{table:classifier_hyperparams}
\end{table}
\end{document}